\documentclass[10pt,twocolumn,letterpaper]{article}

\usepackage{iccv}
\usepackage{times}
\usepackage{epsfig}
\usepackage{graphicx}
\usepackage{amsmath}
\usepackage{amssymb}

\usepackage[pagebackref=true,breaklinks=true,letterpaper=true,colorlinks,bookmarks=false]{hyperref}

\iccvfinalcopy 

\def\iccvPaperID{14} 

\ificcvfinal\fi

\begin{document}

\title{Prose for a Painting}

\author{Prerna Kashyap\\
Columbia University \\
pk2600@columbia.edu
\and
Samrat Phatale \\
Columbia University \\
shp2139@columbia.edu
\and
Iddo Drori \\
Columbia University \\
idrori@cs.columbia.edu
}
\maketitle
\ificcvfinal\thispagestyle{empty}\fi

\begin{abstract}
   Painting captions are often dry and simplistic which motivates us to describe a painting creatively in the style of Shakespearean prose. This is a difficult problem, since there does not exist a large supervised dataset from paintings to Shakespearean prose. Our solution is to use an intermediate English poem description of the painting and then apply language style transfer which results in Shakespearean prose describing the painting. We rate our results by human evaluation on a Likert scale, and evaluate the quality of language style transfer using BLEU score as a function of prose length. We demonstrate the applicability and limitations of our approach by generating Shakespearean prose for famous paintings. We make our models and code publicly available.
\end{abstract}

\section{Introduction}
\begin{figure}
    \includegraphics[width=1\linewidth]{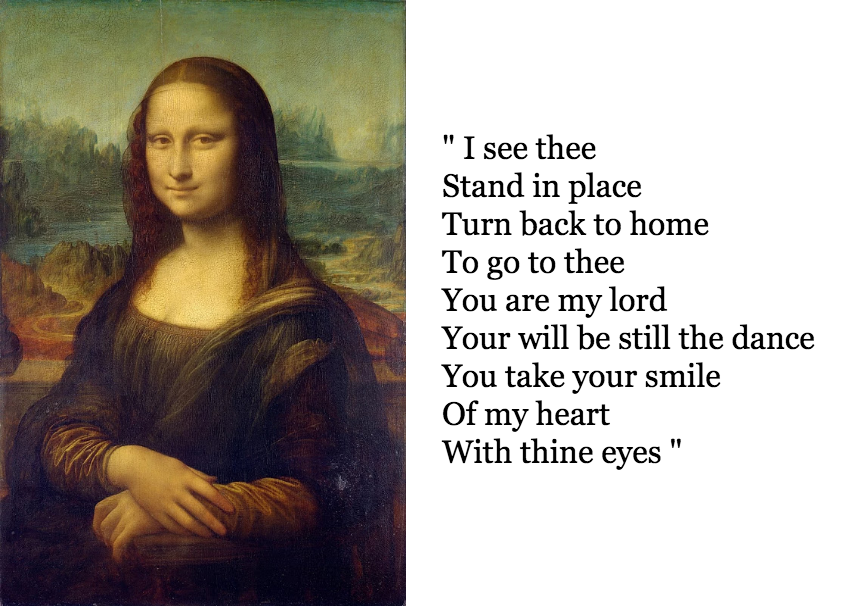}
    \caption{Sample result: Input painting and output synthesized Shakespeare prose.}
    \label{fig:MonaLisa}
\end{figure}

\begin{figure}
    \includegraphics[width=1\linewidth]{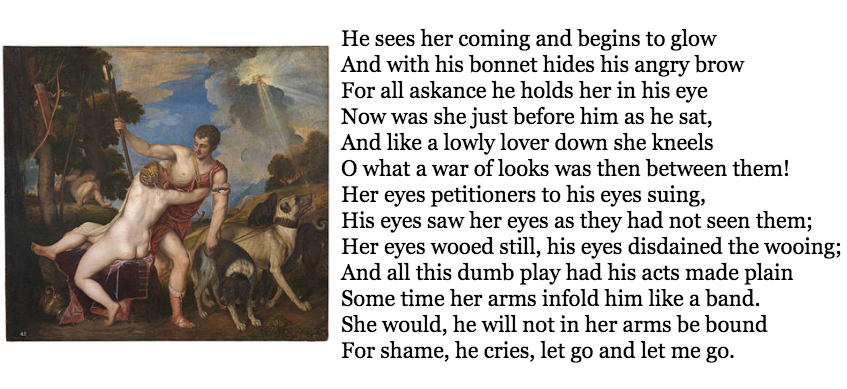}
    \caption{Shakespeare's ground truth prose for the painting Venus and Adonis by Titian.}
    \label{fig:venusandadonis}
\end{figure}

Neural networks have been successfully used to describe images
with text using sequence-to-sequence models 
\cite{vinyals2015show}. However, the results are simple and 
dry captions which are one or two phrases long. Humans looking 
at a painting see more than just objects. Paintings stimulate 
sentiments, metaphors and stories as well. Therefore, our goal 
is to have a neural network describe the painting artistically in
a style of choice. As a proof of concept, we present a model 
which generates Shakespearean prose for a given painting as 
shown in Figure \ref{fig:MonaLisa}. Accomplishing this task is 
difficult with traditional sequence to sequence models since 
there does not exist a large collection of Shakespearean prose 
which describes paintings: Shakespeare's works describes a 
single painting shown in Figure \ref{fig:venusandadonis}. 
Fortunately we have a dataset of modern English poems which 
describe images \cite{liu2018beyond} and line-by-line modern 
paraphrases of Shakespeare's plays 
\cite{jhamtani2017shakespearizing}. Our solution is therefore 
to combine two separately trained models to synthesize Shakespearean prose for a given painting.

\subsection{Related work}
A general end-to-end approach to sequence learning \cite{sutskever2014sequence} makes minimal assumptions on the sequence structure. This model is widely used in tasks such as machine translation, text summarization, conversational modeling, and image captioning. A generative model using a deep recurrent architecture \cite{vinyals2015show} has also beeen used for generating phrases describing an image. The task of synthesizing multiple lines of poetry for a given image \cite{liu2018beyond} is accomplished by extracting poetic clues from images. Given the context image, the network associates image attributes with poetic descriptions using a convolutional neural net. The poem is generated using a recurrent neural net which is trained using multi-adversarial training via policy gradient.

Transforming text from modern English to Shakespearean English using text "style transfer" is challenging. An end to end approach using a sequence-to-sequence model over a parallel text corpus \cite{jhamtani2017shakespearizing} has been proposed based on machine translation. In the absence of a parallel text corpus, generative adversarial networks (GANs) have been used, which simultaneously train two models: a generative model which captures the data distribution, and a discriminative model which evaluates the performance of the generator. Using a target domain language model as a discriminator has also been employed \cite{yang2018unsupervised}, providing richer and more stable token-level feedback during the learning process. A key challenge in both image and text style transfer is separating content from style \cite{drori2003example, gatys2016image, shen2017style}. Cross-aligned auto-encoder models have focused on style transfer using non-parallel text \cite{shen2017style}. Recently, a fine grained model for text style transfer has been proposed \cite{lample2019multiple} which controls several factors of variation in textual data by using back-translation. This allows control over multiple attributes, such as gender and sentiment, and fine-grained control over the trade-off between content and style.

\section{Methods}
\begin{figure*}
    \centering
    \includegraphics[width=0.7\linewidth]{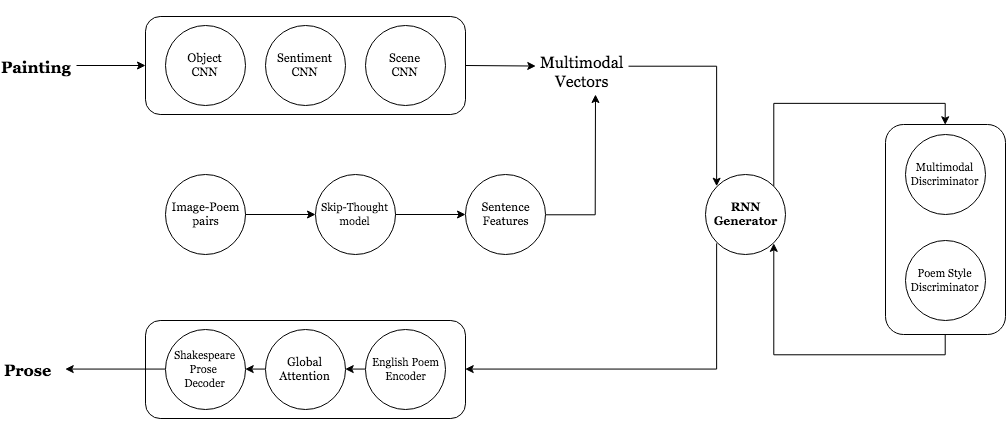}
    \caption{Model architecture: Input is a painting (top left) for which we first generate an English poem using 3 CNNs for feature extraction, and an RNN generator as an agent with 2 discriminators (center right) which provide feedback to the agent for model improvement. The style transfer model takes as input the generator output and performs text style transfer using a seq2seq model with global attention to synthesis the output prose (bottom left).}
    \label{fig:model_architecture}
\end{figure*}

We use a total three datasets: two datasets for generating an English poem from an image, and Shakespeare plays and their English translations for text style transfer. 

We train a model for generating poems from images based on two datasets \cite{liu2018beyond}. The first dataset consists of image and poem pairs, namely a multi-modal poem dataset (MultiM-Poem), and the second dataset is a large poem corpus, namely a uni-modal poem dataset (UniM-Poem). The image and poem pairs are extended by adding the nearest three neighbor poems from the poem corpus without redundancy, and an extended image and poem pair dataset is constructed and denoted as MultiM-Poem(Ex)\cite{liu2018beyond}.

We use a collection of line-by-line modern paraphrases for 16 of Shakespeare’s plays \cite{jhamtani2017shakespearizing}, for training a style transfer network from English poems to Shakespearean prose. We use 18,395 sentences from the training data split. We keep 1,218 sentences in the validation data set and 1,462 sentences in our test set.

\subsection{Image To Poem Actor-Critic Model}
For generating a poem from images we use an existing actor-critic architecture \cite{liu2018beyond}. This involves 3 parallel CNNs: an object CNN, sentiment CNN, and scene CNN, for feature extraction. These features are combined with a skip-thought model which provides poetic clues, which are then fed into a sequence-to-sequence model trained by policy gradient with 2 discriminator networks for rewards. This as a whole forms a pipeline that takes in an image and outputs a poem as shown on the top left of Figure \ref{fig:model_architecture}. A CNN-RNN generative model acts as an agent. The parameters of this agent define a policy whose execution determines which word is selected as an action. When the agent selects all words in a poem, it receives a reward. Two discriminative networks, shown on the top right of Figure \ref{fig:model_architecture}, are defined to serve as rewards concerning whether the generated poem properly describes the input image and whether the generated poem is poetic. The goal of the poem generation model is to generate a sequence of words as a poem for an image to maximize the expected return.

\subsection{Shakespearizing Poetic Captions}
For Shakespearizing modern English texts, we experimented with various types of sequence to sequence models. Since the size of the parallel translation data available is small, we leverage a dictionary providing a mapping between Shakespearean words and modern English words to retrofit pre-trained word embeddings. Incorporating this extra information improves the translation task. The large number of shared word types between the source and target sentences indicates that sharing the representation between them is beneficial.

\subsubsection{Seq2Seq with Attention}
We use a sequence-to-sequence model which consists of a single layer unidrectional LSTM encoder and a single layer LSTM decoder and pre-trained retrofitted word embeddings shared between source and target sentences. We experimented with two different types of attention: global attention \cite{luong2015effective}, in which the model makes use of the output from the encoder and decoder for the current time step only, and Bahdanau attention \cite{bahdanau2015neural}, where computing attention requires the output of the decoder from the prior time step. We found that global attention performs better in practice for our task of text style transfer.

\subsubsection{Seq2Seq with a Pointer Network}
Since a pair of corresponding Shakespeare and modern English sentences have significant vocabulary overlap we extend the sequence-to-sequence model mentioned above using pointer networks \cite{merity2016pointer} that provide location based attention and have been used to enable copying of tokens directly from the input. Moreover, there are lot of proper nouns and rare words which might not be predicted by a vanilla sequence to sequence model.


\subsubsection{Prediction}
For both seq2seq models, we use the attention matrices returned at each decoder time step during inference, to compute the next word in the translated sequence if the decoder output at the current time step is the UNK token. We replace the UNKs in the target output with the highest aligned, maximum attention, source word. The seq2seq model with global attention gives the best results with an average target BLEU score of 29.65 on the style transfer dataset, compared with an average target BLEU score of 26.97 using the seq2seq model with pointer networks.

\section{Results}
\begin{figure}
    \includegraphics[width=1\linewidth]{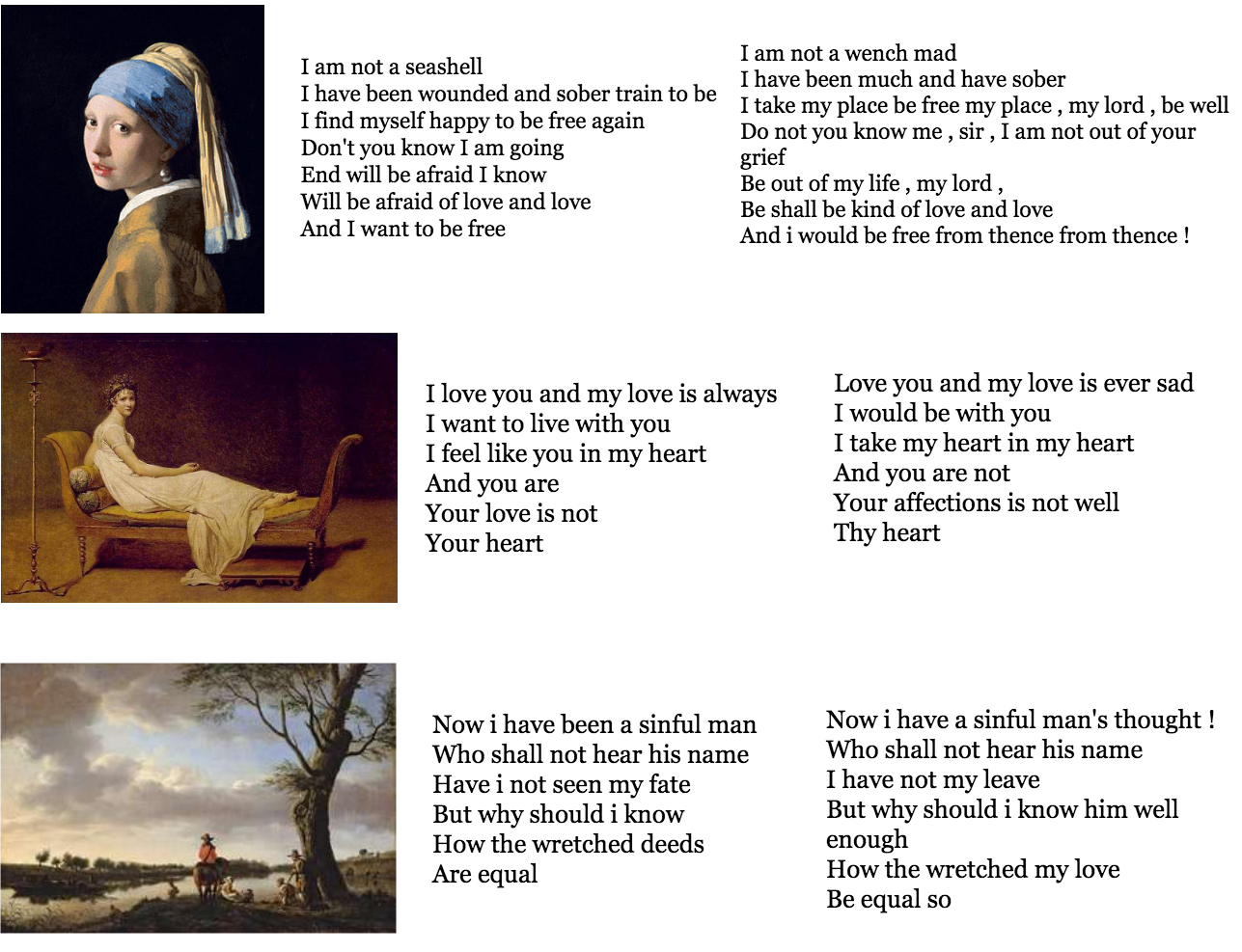}
    \label{fig:examples}
    \caption{Sample results: Painting (left), synthesized English poem (center) and Shakespearean Prose (right) for "Girl with the pearl earing", a painting of a man on a lake, and "Portrait of Madame Recaimer".}
    \label{fig:examples}
\end{figure}

\begin{figure}
    \centering
    \includegraphics[width=0.8\linewidth]{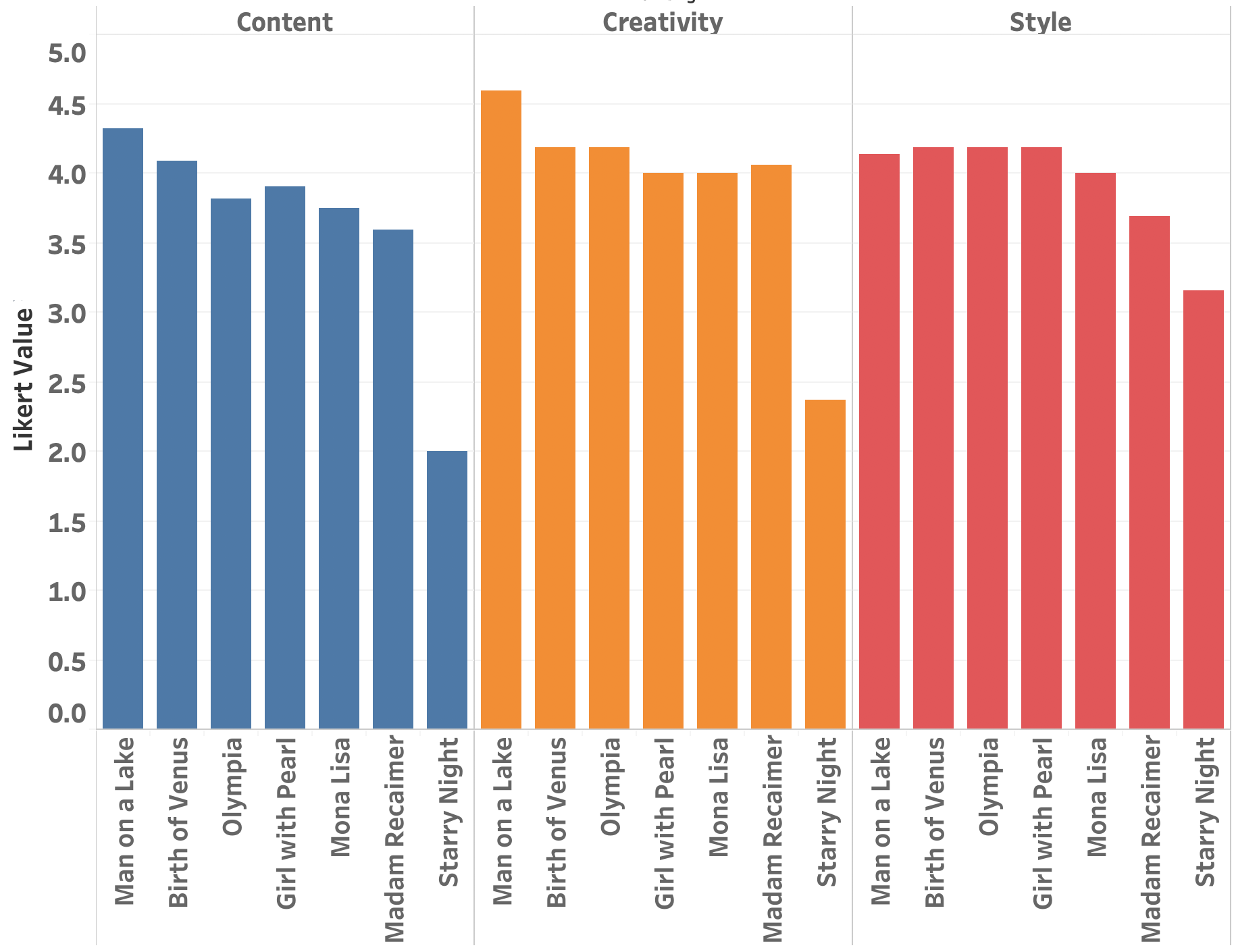}
    \caption{Survey results: Average rating of content, creativity, and style on a Likert scale.}
    \label{fig:survey}
\end{figure}

We perform a qualitative analysis of the Shakespearean prose generated for the input paintings. We conducted a survey, in which we presented famous paintings including those shown in Figures \ref{fig:MonaLisa} and \ref{fig:examples} and the corresponding Shakespearean prose generated by the model, and asked 32 students to rate them on the basis of content, creativity and similarity to Shakespearean style on a Likert scale of 1-5. Figure \ref{fig:survey} shows the result of our human evaluation.

The average content score across the paintings is 3.7 which demonstrates that the prose generated is relevant to the painting. The average creativity score is 3.9 which demonstrates that the model captures more than basic objects in the painting successfully using poetic clues in the scene. The average style score is 3.9 which demonstrates that the prose generated is perceived to be in the style of Shakespeare. 

\begin{figure}
    \includegraphics[width=0.8\linewidth]{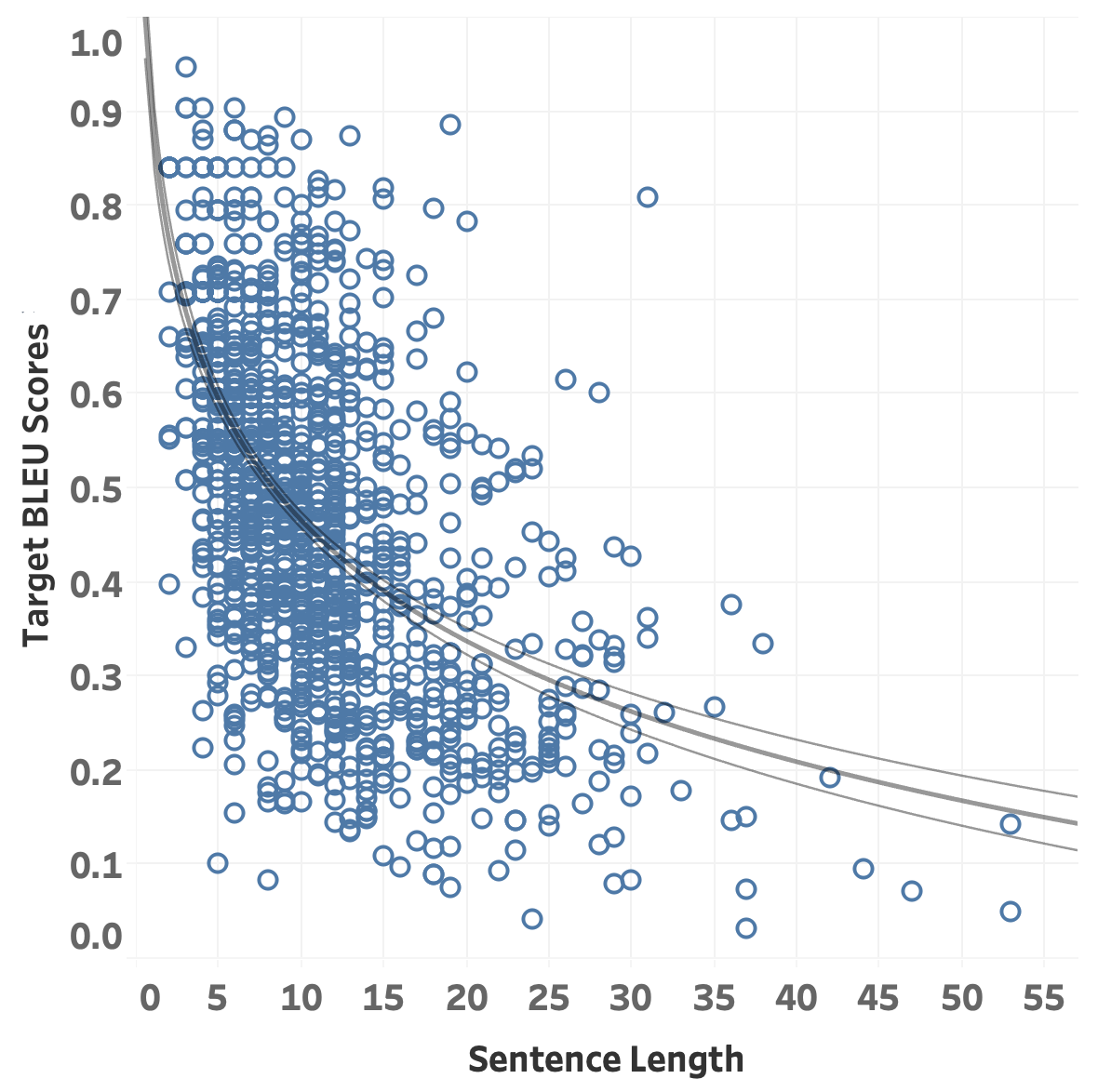}
    \caption{BLEU score vs. source sentence length.}
    \label{fig:bleuplot}
\end{figure}

We also perform a quantitative analysis of style transfer by generating BLEU scores for the model output using the style transfer dataset. The variation of the BLEU scores with the source sentence lengths is shown in Figure \ref{fig:bleuplot}. As expected, the BLEU scores decrease with increase in source sentence lengths.


\subsection{Implementation}
All models were trained on Google Colab with a single GPU using Python 3.6 and Tensorflow 2.0. The number of hidden units for the encoder and decoder is 1,576 and 256 for seq2seq with global attention and seq2seq with pointer networks respectively. Adam optimizer was used with the default learning rate of 0.001. The model was trained for 25 epochs. We use pre-trained retrofitted word embeddings of dimension 192.

\subsection{Limitations}
Since we do not have an end-to-end dataset, the generated English poem may not work well with Shakespeare style transfer as shown in Figure \ref{fig:survey} for "Starry Night" with a low average content score. This happens when the style transfer dataset does not have similar words in the training set of sentences. A solution would be to expand the style transfer dataset, for a better representation of the poem data.


\subsection{Conclusions and Future Work}
In conclusion, combining two pipelines with an intermediate representation works well in practice. We observe that a CNN-RNN based image-to-poem net combined with a seq2seq model with parallel text corpus for text style transfer synthesizes Shakespeare-style prose for a given painting. For the seq2seq model used, we observe that it performs better in practice using global attention as compared with local attention. We make our models and code publicly available  \cite{proseforpaintingcode}.
In future work we would like to experiment with GANs in the absence of non-parallel datasets, so that we can use varied styles for text style transfer. We would also like to experiment with cross aligned auto-encoders, which form a latent content representation, to efficiently separate style and content.

{\small
\bibliographystyle{ieee_fullname}
\bibliography{bibliography}
}

\end{document}